\documentclass[runningheads]{llncs}

\usepackage{graphicx}
\usepackage{amsmath}
\begin{document}
\title{Joint segmentation and discontinuity-preserving deformable registration: Application to cardiac cine-MR images}
%
\author{Xiang Chen\inst{1} \and 
Yan Xia\inst{1,2} \and 
Nishant Ravikumar\inst{1,2} \and 
Alejandro F Frangi\inst{1,2,3,4,5}} 
\authorrunning{X. Chen et al.}
%
\institute{Center for Computational Imaging and Simulation Technologies in Biomedicine, School of Computing, University of Leeds, Leeds,UK \and
Biomedical Imaging Department, Leeds Institute for Cardiovascular and Metabolic Medicine, School of Medicine University of Leeds, Leeds, UK\\
\and Department of Cardiovascular Sciences, KU Leuven, Leuven, Belgium\\
\and Department of Electrical Engineering, KU Leuven, Leuven, Belgium
\and Alan Turing Institute, London, UK}

%
\maketitle              
\sloppy
\begin{abstract}
Medical image registration is a challenging task involving the estimation of spatial transformations to establish anatomical correspondence between pairs or groups of images. Recently, deep learning-based image registration methods have been widely explored, and demonstrated to enable fast and accurate image registration in a variety of applications. However, most deep learning-based registration methods assume that the deformation fields are smooth and continuous everywhere in the image domain, which is not always true, especially when registering images whose fields of view contain discontinuities at tissue/organ boundaries. In such scenarios, enforcing smooth, globally continuous deformation fields leads to incorrect/implausible registration results. We propose a novel discontinuity-preserving image registration method to tackle this challenge, which ensures globally discontinuous and locally smooth deformation fields, leading to more accurate and realistic registration results. The proposed method leverages the complementary nature of image segmentation and registration and enables joint segmentation and pair-wise registration of images. A co-attention block is proposed in the segmentation component of the network to learn the structural correlations in the input images, while a discontinuity-preserving registration strategy is employed in the registration component of the network to ensure plausibility in the estimated deformation fields at tissue/organ interfaces. We evaluate our method on the task of intra-subject spatio-temporal image registration using large-scale cinematic cardiac magnetic resonance image sequences, and demonstrate that our method achieves significant improvements over the state-of-the-art for medical image registration, and produces high-quality segmentation masks for the regions of interest. 

\end{abstract}

\section{Introduction}
Image registration involves establishing spatial correspondence between a given pair or group of images, which is fundamental for many downstream medical imaging applications (e.g. image fusion, atlas-based segmentation, image-guided interventions, organ motion tracking and strain analysis, amongst others). Recently, deep learning-based methods have found widespread use in medical image registration, achieving comparable or better performance than traditional registration methods, and yielding substantial speed-ups in execution relative to the latter. Among them, unsupervised and weakly-supervised methods are the most popular as they do not require ground-truth deformation fields to be available. Unsupervised methods~\cite{vos2017end,balakrishnan2018unsupervised,dalca2018unsupervised} do not need any ground-truth annotations (e.g. segmentation, landmarks and ground-truth deformation fields) for training and rely just on the information available in the pair/group of images to be registered. These approaches have been shown to achieve similar or better registration performance than traditional registration methods~\cite{chen2021deep}, at a fraction of the execution time. Weakly-supervised methods~\cite{balakrishnan2019voxelmorph,dalca2019unsupervised}, however, require some annotations (e.g. landmarks, segmentation masks) for training, but have been shown to improve registration performance relative to their unsupervised counterparts.

Currently, most deep learning-based registration methods assume globally smooth and continuous deformation fields throughout the image domain, using regularisation like L2 norm of deformation fields to ensure that. However, this assumption is not appropriate for all medical image registration applications, especially when there are physical discontinuities resulting in sliding motion between organs/soft tissues that must be estimated to register the input images. For example, respiratory motion resulting from inflation and deflation of the lungs during breathing contains discontinuities between the lungs, the pleural sac encompassing the lungs and the surrounding rib cage. The pleural sac itself contains two layers that slide over one another as the lung inflates and deflates resulting in what is perceived as a sliding motion at the boundaries of the lungs. Enforcing deformation fields to be completely smooth when registering thoracic images of any given individual to recover breathing motion would result in physically unrealistic deformation fields and artefacts near lung boundaries.

Generally, the sliding of organs and different material properties of different sub-regions in the input images may cause the deformation fields to be locally smooth but globally discontinuous~\cite{hua2016non}. In previous research, many traditional methods have been proposed to achieve discontinuity-preserving image registration~\cite{hua2016non,wu2008evaluation,schmidt2012estimation,pace2013locally,hua2017multiresolution,demirovic2015bilateral,vishnevskiy2016isotropic,nie2019deformable,li2018discontinuity,zhang2016improved}. The fundamental goal of discontinuity-preserving registration is to predict deformation fields which are locally smooth, i.e. within each sub-region, while, discontinuous globally, such as at the interface between different regions/organs.
A simple solution to this problem is to register the corresponding sub-organs in the input images independently and then compose them to obtain the final deformation field~\cite{wu2008evaluation}. Another approach is to reformulate the regularisation constraint enforced on the estimated deformation field/functions, to allow for discontinuities at points near the interface between different tissues/organs~\cite{schmidt2012estimation,pace2013locally}. However, for those methods, the label information (segmentation masks/landmark) is generally required to delineate where the discontinuity may occur, which may not always be available in realistic scenarios. Therefore, some research has explored achieving discontinuity-preserving registration without requiring \textit{a priori} definition of segmentation masks/landmarks, like using vectorial total variation regularisation~\cite{vishnevskiy2016isotropic} or bounded formation theory~\cite{nie2019deformable}.

Most existing deep learning-based image registration methods do not tackle the problem of estimating deformation fields that preserve discontinuities at tissue/organ boundaries and generally regularise the estimated deformations to be globally smooth and continuous across the image domain. Ng et al.\cite{ng2020unsupervised} was the first to propose a custom discontinuity-preserving regulariser to constrain the estimation of deformation fields and guide the training of a deep registration network. They assumed that the motion vectors should be parallel to each other, and achieved it by minimising the unsigned area of the parallelogram spanned by two displacement vectors associated with moving image voxels. The advantage of this method was that it did not require label information like segmentation/landmarks. However, it was unable to locate the accurate position of discontinuity that may occur and thereby did not show significant improvement than traditional methods. Previously, we proposed a deep neural network ~\cite{chen2021ddir} to register pairs of corresponding anatomical structures in the input images separately, and compose the deformation fields of each pair to obtain the final deformation field used to warp the source/moving image to the target/fixed image. Instead of using a globally smooth regularisation, the smooth regularisation is applied to each sub-deformation field, which ensures the final deformation fields are locally smooth while globally discontinuous. The proposed approach is shown to significantly overwhelm state of the art traditional and deep learning based registration methods. However, it requires segmentation masks to split both the moving and fixed images into corresponding pairs of anatomical regions/structures, during both training and testing, which limits its utility in real scenarios (e.g. segmentation masks may not be readily available for the regions/structures of interest and trained segmentation models to supplement the same may not available either).

This paper is an extension of our previous work, namely, the deep discontinuity-preserving registration method (DDIR) presented at the Medical Image Computing and Computer Assisted Intervention (MICCAI) 2021 conference~\cite{chen2021ddir}. In~\cite{chen2021ddir}, the segmentation masks were required during both training and testing to split the original moving and fixed images into pairs of corresponding anatomical regions, limiting its application to scenarios where segmentation masks are readily available for the anatomical regions of interest or can be predicted automatically using a suitable segmentation approach. In this paper, we propose a joint registration and segmentation approach wherein, a segmentation module is incorporated within DDIR, which we refer to as SDDIR. This ameliorates the need for segmenting the fixed and moving images prior to registering them. Instead of using ground-truth segmentation, SDDIR applies the predicted segmentation masks to split the moving and fixed images into corresponding pairs of anatomical regions/structures. A co-attention block is used within the segmentation module to learn the structural correlation between the moving and fixed images, and further improve the segmentation performance. A comprehensive set of experiments conducted using a large-scale cardiac cinematic magnetic resonance (cine-MR) imaging dataset, available in the UK biobank (UKBB) study~\cite{petersen2015uk}, is used to demonstrate that the proposed approach outperforms competing methods in terms of registration accuracy, whilst also yielding high-quality segmentation masks of the cardiac structures of interest in the fixed and moving images. Additionally, we also demonstrate the generalisation of our method by transferring the pre-trained network on UKBB to two external cardiac MR image datasets, Automatic Cardiac Diagnosis Challenge (ACDC~\cite{bernard2018deep}) and Multi-Centre, Multi-Vendor \& Multi-Disease Cardiac Image Segmentation Challenge~\cite{campello2021multi} (M\&M).

\subsection{Related Work}
\subsubsection{Jointly Segmentation and Registration}
Image segmentation and image registration are both fundamental tasks in computer vision and medical image analysis, which share similarities with each other in terms of the visual cues/features that are relevant for solving either task. Some traditional methods have considered these two independent tasks simultaneously, leading to improved performance in both tasks~\cite{droske2007multiscale,dong2017scalable}, due to their complementary nature. For example, Droske et al.~\cite{droske2007multiscale} presented an variational approach to achieve multi-tasks: the detection of corresponding edges, edge-preserving denoising, and morphological registration. They demonstrated that the edge detection and registration tasks were beneficial to the other, with the local weak edge detection improving registration performance and vice versa. Similarly, Dong et al.~\cite{dong2017scalable} designed a joint segmentation and registration method for infant brain image registration and found the segmentation and registration steps were mutually beneficial.

In deep-learning based methods, several previous research have proposed to achieve segmentation by registration (atlas-based segmentation~\cite{sinclair2022atlas}) or use segmentation to improve registration (e.g. weakly-supervised registration~\cite{chen2021ddir}). Most recently, studies have proposed to tackle these tasks jointly in a single end-to-end framework~\cite{xu2019deepatlas,li2019hybrid,qiu2021u}, which generally includes two parallel sub-networks, the segmentation sub-network and registration sub-network. A composite loss function is used to train these networks, comprising four terms, namely, the intensity similarity loss, the segmentation loss on moving/fixed images, the regularisation term enforcing estimated deformation fields to be globally smooth, and a segmentation consistency loss term. The segmentation consistency loss is computed on warped moving segmentation and ground-truth fixed segmentation, and is shared by both sub-networks, which helps to improve both segmentation and registration performance.
In~\cite{xu2019deepatlas}, Xu et al. proposed a novel joint segmentation and registration network, named DeepAtlas, which was flexible and could be applied in samples without label segmentation (segmentation masks). In the training stage, the registration sub-network and segmentation sub-network are trained alternately, with the ground-truth segmentation missing in some of the training samples. To train the registration sub-network, when the ground-truth segmentation was available, the segmentation consistency loss was computed based on the ground-truth segmentation masks, otherwise using the predicted moving and fixed segmentation from the segmentation sub-network. They demonstrated that their method could achieve significant improvement in the segmentation and registration than sole segmentation or registration networks. Different from~\cite{xu2019deepatlas}, Li et al.~\cite{li2019hybrid} trained both sub-networks simultaneously, with the same loss function. They only segmented the moving image in the network and used segmentation accuracy (computed between the predicted moving segmentation and the ground truth moving segmentation mask) and segmentation consistency (computed between the warped moving segmentation and the fixed segmentation mask) for network training. Subsequent studies have also explored removing the requirements of ground-truth segmentation on the segmentation part based on Bayesian inference with probabilistic atlas~\cite{qiu2021u}, or extending the idea of jointly learning segmentation and registration to multi-modal image registration~\cite{chen2021mr}.

Segmentation sub-networks in existing joint segmentation and registration approaches generally segment fixed and moving images independently (or only segment the moving images), and ignore the inherent correlations that exist between them. To exploit this correlated structural information, with a view to enhance joint segmentation and registration performance, we employ a `co-attention' based segmentation sub-network within the proposed approach to jointly segment the fixed and moving input images.

\subsubsection{Co-attention based Segmentation}
Co-attention based segmentation aims to improve segmentation performance by sufficiently leveraging the structural correlations that exist between multiple input images to be segmented. Generally, there are at least two inputs images, which contain the same type of objects to segment. By learning common/correlated features from multiple images containing the same objects, the co-attention block has been shown to improve segmentation robustness and accuracy for the objects of interest~\cite{li2019group,ahn2021multi,mo2021mutual,chen2020show,chen2018semantic,lei2020self}.

The co-attention block is generally used to automatically establish correspondence between correlated regions in input images/feature representations through training on large-scale data, where, the correlated regions would be enhanced while other parts of the images are suppressed. The most popular type of co-attention is spatial co-attention~\cite{ahn2021multi,chen2020show,yang2021learning}, where the co-attention establishes correspondence within the spatial domain of the input images. Sometimes an additional channel co-attention is also used prior to the spatial co-attention~\cite{li2019group,chen2018semantic,lei2020self}. Spatial co-attention has been predominantly applied to image features, however, recent studies have also applied it to graph features learned in graph neural networks~\cite{mo2021mutual}. 
Li et al.~\cite{li2019group} utilised co-attention within a recurrent neural network architecture to learn correlated structural information across a group of images and improve segmentation performance by suppressing the influence of uncorrelated/noisy information. In the group-wise training objective, they used the cross-image similarity between the co-occurring objects and figure-ground distinctness (i.e. distinctness between the detected co-occurring objects and the rest of the images like background) as additional supervision. 
Additionally, co-attention has also been used for the segmentation of the same object in different time frames, in a video sequence for example. Ahn et al.~\cite{ahn2021multi} proposed a Multi-frame Attention Network to learn highly correlated spatio-temporal features in a sequence. Experiments demonstrated that their method significantly outperformed other competing deep learning-based methods. Furthermore, Yang~\cite{yang2021learning} proposed a zero-shot object detection approach for analysing video sequences, using co-attention to learn motion patterns. They empirically demonstrated that their approach outperformed previous zero-shot video object segmentation approaches, while requiring fewer training data.

In this study, we tackle the problem of intra-subject registration of cardiac cine-MR images, acquired at different phases/time points in the cardiac cycle. 
In intra-subject cardiac image registration the moving and fixed images are different/deformed representations of the same heart, acquired at different time points in the cardiac cycle (e.g. at end-diastole (ED) to end-systole (ES)). Hence, we hypothesise that joint segmentation of both the fixed and moving images using co-attention can yield more consistent segmentations of the cardiac regions/structures of interest, and in turn improve the overall registration performance of the proposed approach.

\subsubsection{Discontinuity-preserving Image Registration}
Discontinuity-preserving image registration has been widely explored in traditional iterative optimisation based registration approaches but remains relatively unexplored in the context of deep learning-based image registration. In medical image registration, due to varying material properties properties between different tissues/organs, and the physical discontinuities that exist at their boundaries, the underlying deformations that must be recovered to register the images are often locally smooth and globally discontinuous (e.g. at the boundaries between different organs~\cite{hua2016non}). Consequently, registration methods which constrain estimated deformation fields to be globally smooth, lead to implausible deformations at tissue/organ boundaries. Traditional discontinuity-preserving registration methods can be roughly divided into two categories, those that use additional weak labels such as contours/segmentation masks to guide image registration and others that do not. Methods that require segmentation masks or contour key points delineating the discontinuities of interest between structures/organs in the images, can be further summarised into two categories - (1) registering different sub-regions in the images independently~\cite{wu2008evaluation,risser2013piecewise,von20074d}; (2) using custom regularisation constraints to preserve global discontinuities~\cite{schmidt2012estimation,pace2013locally,jud2017directional} or revising the interpolation function~\cite{hua2016non,hua2017multiresolution} at boundaries between different image sub-regions. These methods have been demonstrated to generate more realistic deformation fields and achieve more accurate results than registration methods that assume globally smooth deformation fields~\cite{wu2008evaluation,schmidt2012estimation,pace2013locally,hua2017multiresolution}.

The aforementioned approaches generally require segmentation masks/landmarks delineating the boundaries of objects/structures of interest, which are not always readily available. Several approaches addressed this issue by revising the regularisation term in the loss/energy function~\cite{demirovic2015bilateral,sandkuhler2018adaptive,vishnevskiy2016isotropic,nie2019deformable,li2018discontinuity,zhang2016improved}, either using some label information like segmentation masks or contour points/landmarks computed pre-registration, or based on different assumptions for physical property of the deformation fields (e.g. isotropic total variation or bounded deformation). 
Li et al.~\cite{li2018discontinuity} designed a two-stage registration framework to tackle this issue. They predicted a coarse segmentation mask based on the motion fields predicted during the first stage of image registration, which used mask-free regularisation. Subsequently, in the second stage, the smoothness constraint was relaxed at object/structure boundaries with discontinuities, using masked regularisation and masked interpolation. Similarly, Sandkuhler~\cite{sandkuhler2018adaptive} proposed an adaptive edge weight function based on local image intensities and transformation fields to detect the sliding organ boundaries, then applied an adaptive anisotropic graph diffusion regularisation in the Demons registration to achieve discontinuity-preserving image registration. 
Some previous approaches do not need to compute any weak label information, prior to registering images~\cite{demirovic2015bilateral,vishnevskiy2016isotropic,nie2019deformable,nie2021deformable}.
Demirovic et al.~\cite{demirovic2015bilateral} proposed to replace the Gaussian filter of the accelerated Demons with a bilateral filter, using information from both displacement and image intensity. By adjusting two tunable parameters, they could obtain more realistic deformations in the presence of discontinuities. 
Vishnevskiy et al.~\cite{vishnevskiy2016isotropic} designed an isotropic total variation regularisation approach for B-splines based image registration, to enable non-smooth deformation fields and used the Alternating Directions Method of Multipliers to solve it. Their method did not require organ masks and could estimate the motion of organs/structures either side of the discontinuous boundary separating them. By assuming the desired deformation field to be a function of bounded deformation/bounded generalized deformation (referring to~\cite{temam1980functions}), Nie et al~\cite{nie2019deformable,nie2021deformable} built novel variational frameworks to allow possible discontinuities of displacement fields in images, outperforming~\cite{vishnevskiy2016isotropic}.

Most deep learning-based image registration methods assume the desired deformation fields to be globally smooth and continuous, and do not consider the presence or relevance of discontinuities at structure/organ boundaries and their impact on the image registration task. To our best knowledge, only two previous studies have attempted to preserve discontinuities at object/structure boundaries in deep learning-based image registration~\cite{ng2020unsupervised,chen2021ddir}. Ng et al.~\cite{ng2020unsupervised} addressed this issue in an unsupervised manner. They proposed an discontinuity preserving regularisation term by comparing local displacement vectors with neighboring displacement vectors individually, which was able to tackle specific behaviors on the discontinues deformation fields. Without ground-truth information specifying the locations of discontinuous boundaries, their registration performance did not show significant improvements than traditional approaches. In contrast, we previously~\cite{chen2021ddir} proposed a deep discontinuity-preserving image registration (DDIR) approach, to generate locally smooth sub-deformation fields for each image sub-region, which were then composed to obtain locally smooth and globally discontinuous deformation fields. Although \cite{chen2021ddir} was shown to outperform the state of the art, its need for segmentation masks delineating the objects/structures of interest during inference, limits its application in real-world scenarios. Therefore, in this paper, to tackle this issue, a joint segmentation and registration approach is proposed for discontinuity-preserving registration, which only requires ground-truth masks in the training process.

\section{Method}
In this paper, we focus on pair-wise image registration, aiming to establish spatial correspondence between the moving image $\textbf{I}_M$ and fixed image $\textbf{I}_F$. This task can be formulated as,
\begin{equation}
\label{eqn:formula}
\phi(\textbf{x}) = \textbf{x} + u(\textbf{x}),
\end{equation}
where, $\textbf{x}$ is the coordinate of voxels/pixels in the moving image $\textbf{I}_M$, $u(\textbf{x})$ and $\phi(\circ)$ represents the displacement field and the deformation function, respectively.

\begin{figure*}[ht]
\begin{center}
\includegraphics[width=\textwidth]{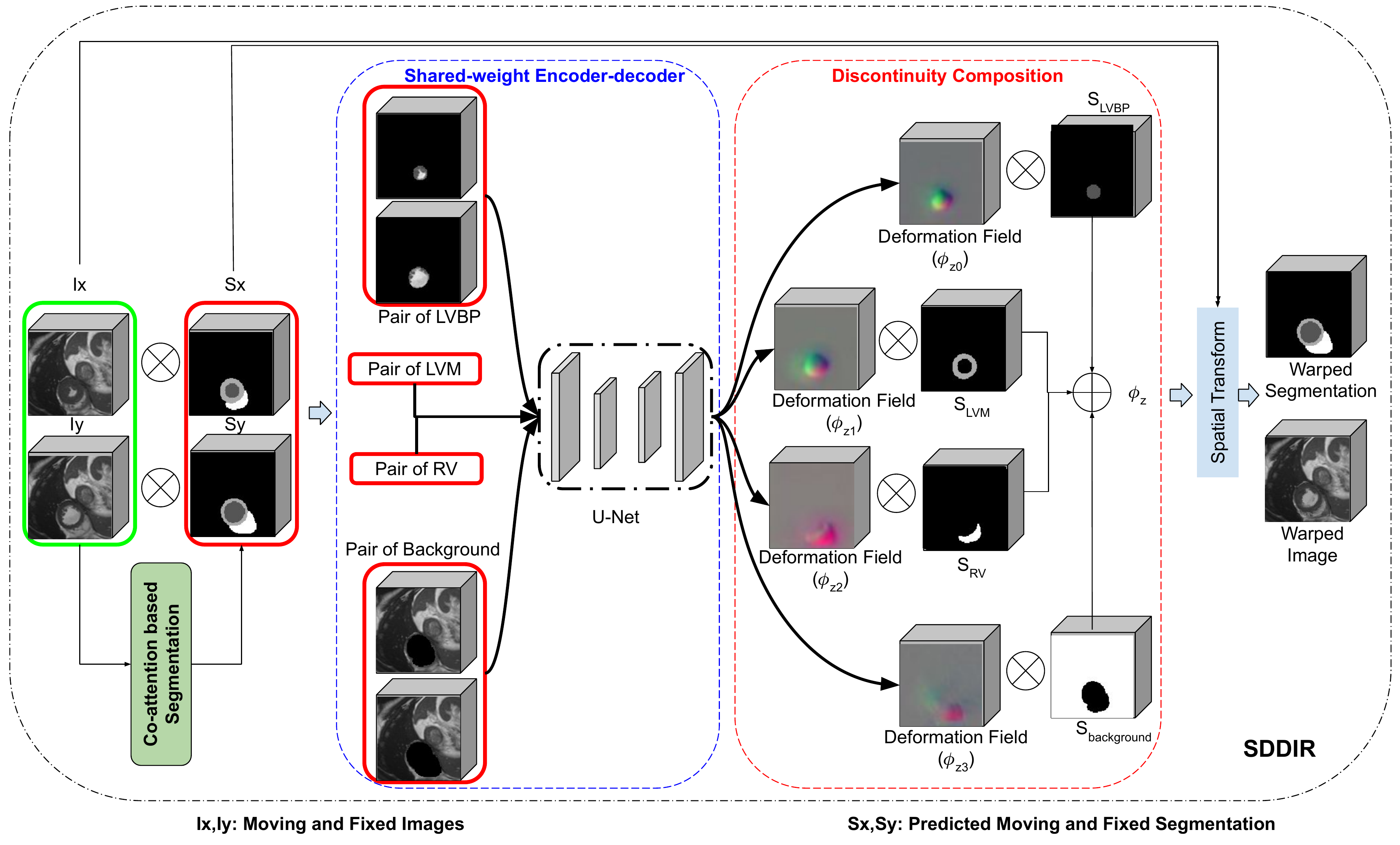}
\caption{Schema of SDDIR. The registration network applies a shared-weight U-Net to extract features from pairs of LVBP, LVM, RV and background. Based on them, we obtain four sub-deformation fields for different regions. The final deformation field is obtained by composing these four deformation fields with corresponding segmentation. The cardiac MR images were reproduced by kind permission of UK Biobank ©.}
\label{fig:network}
\end{center}
\end{figure*}

To preserve discontinuities during image registration, similar to our previous study~\cite{chen2021ddir}, we decompose the fixed and moving images into corresponding pairs of image sub-regions, register each pair and combine the obtained sub-deformation fields to obtain the final deformation field used to warp the moving image. Different from~\cite{chen2021ddir}, in this study we propose a joint segmentation and discontinuity-preserving image registration (SDDIR) approach, which includes an additional segmentation sub-network in DDIR \cite{chen2021ddir}, to jointly segment the fixed and moving images. The primary motivation for the approach proposed in this study is to ameliorate the need for separately sourcing segmentation masks (either manually or automatically) for the images to be registered, as required by DDIR. As shown in Fig.~\ref{fig:network}, our SDDIR includes a segmentation sub-network and a registration sub-network. The input fixed and moving images are first fed into the segmentation branch and tissue/organ specific segmentation masks are predicted for each image. For example, the focus of this study is on intra-subject cardiac cine-MR image registration, and given input cine-MR images, four-class segmentation masks are predicted, delineating the following regions - left ventricular myocardium, left ventricular blood pool, right ventricular blood pool and background tissue. Subsequently, the fixed and moving images, and their predicted segmentation masks are fed into the discontinuity-preserving image registration branch, which predicts region/structure-specific sub-deformation fields, and composes them in to a final deformation field used to warp the moving image. The segmentation and registration branches are trained jointly end-to-end, as a single network, using a combined composite loss function. In subsequent sections, we describe the co-attention based segmentation sub-network, discontinuity-preserving image registration sub-network, and the composite loss function used in the proposed approach.

\subsection{Co-attention Based Segmentation}
\begin{figure*}[ht]
\begin{center}
\includegraphics[width=\textwidth]{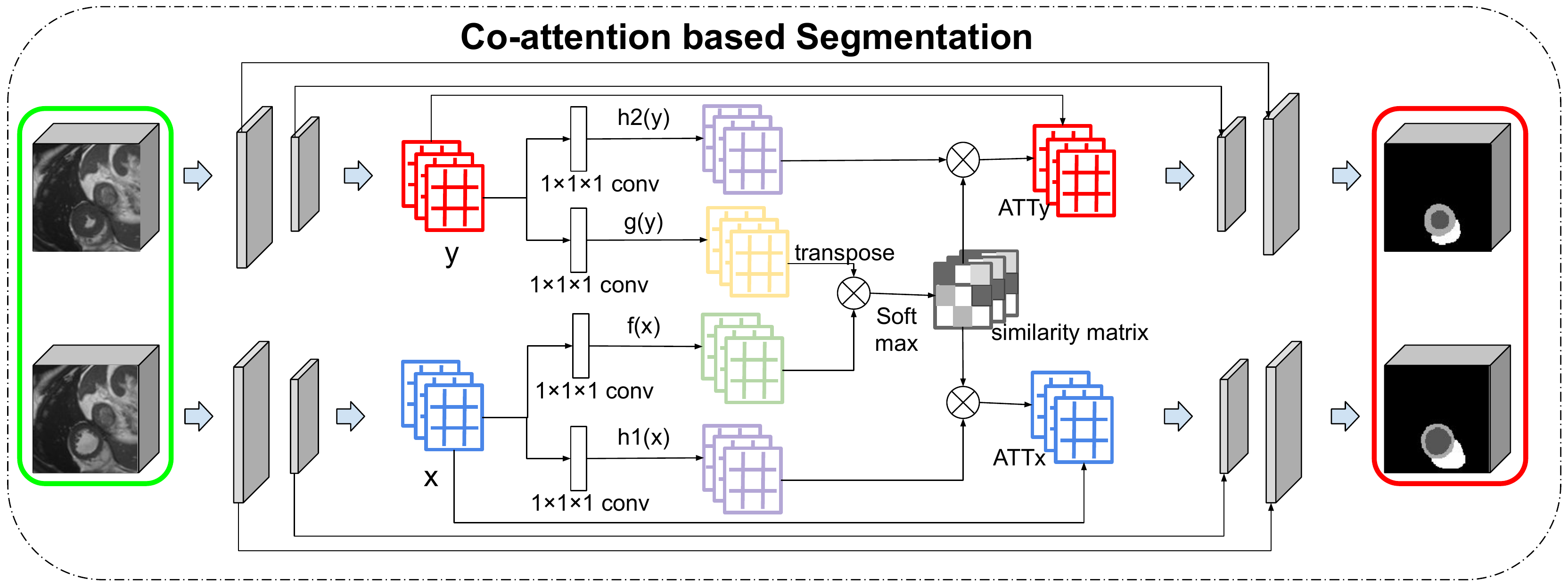}
\caption{Schema of co-attention based segmentation. The x and y denotes the feature of moving and fixed images, respectively. The cardiac MR images were reproduced by kind permission of UK Biobank ©.}
\label{fig:segmentation}
\end{center}
\end{figure*}

The segmentation sub-network in SDDIR is based on a 3D U-Net~\cite{cciccek20163d}, with a co-attention block in the bottleneck layer designed to learn structural correlations between the fixed and moving images, as shown in Fig.~\ref{fig:segmentation}. In the segmentation sub-network, the encoder and decoder branches each comprise two pairs of downsampling and upsampling convolution blocks, respectively. The encoder contains two separate channels to encode the original moving and fixed images from $R^{ H \times W \times D}$ into features of $R^{ \frac{H}{4} \times \frac{W}{4} \times \frac{D}{4}}$. Each encoder channel uses two downsampling blocks (comprising a convolution layer, an activation and an average-pooling layer). The bottleneck layer contains the co-attention block which takes fixed and moving image features extracted by the encoder as inputs and predicts corresponding attention maps for the same. Similarly, in the decoder, two separate channels comprising two upsampling convolution blocks each, are used to predict segmentation masks for the fixed and moving images in their original size/resolution, given their corresponding attention feature maps as inputs. Here, each upsampling block comprises an upsampling layer, a convolution layer and an activation layer. To improve network training and performance, skip-connections are used to concatenate features from each block in the encoder with its corresponding block (i.e. at the same spatial resolution) in the decoder.

The co-attention block predicts task-specific attention feature maps for both the fixed and moving images, where relevant pixels are enhanced while the rest of are suppressed. The moving and fixed image feature maps $\textbf{F}_{mov},\textbf{F}_{fix} \in \mathcal{R}^{ W \times H \times D \times C}$ (C, W, H and D are channels, width, height and depth of feature maps, respectively) are first transformed into two different feature spaces (denoted as $f(\circ)$ and $g(\circ)$) by two $1\times1\times1$ convolution layers and flattened (from $\mathcal{R}^{W \times H \times D \times C}$ to $\mathcal{R}^{N\times C}, N=W \times H \times D$) to calculate similarity matrix $\textbf{S} \in \mathcal{R}^{N\times N}$. With the similarity matrix $\textbf{S}$ and the feature maps learned from the inputs using two additional $1\times1\times1$ convolution layers (denoted $h_1(\circ)$ and $h_2(\circ)$), the fixed attention maps $\textbf{ATT}_{fix}$ and moving attention maps $\textbf{ATT}_{mov}$ are computed. The process of co-attention can be formulated as, 
\begin{equation}
\label{eqn:co-attention}
\begin{aligned}
& \textbf{S} = f(\textbf{F}_{mov}) \times g(\textbf{F}_{fix})^T,\\
& \textbf{ATT}_{mov} = Softmax(\textbf{S}) \times h_2(\textbf{F}_{fix}),\\
& \textbf{ATT}_{fix} = Softmax(\textbf{S}^T) \times h_1(\textbf{F}_{mov}),\\
& \textbf{O}_{mov} = Concat(\textbf{F}_{mov}, \sigma(\textbf{ATT}_{mov}) \cdot \textbf{ATT}_{mov}), \\
& \textbf{O}_{fix} = Concat(\textbf{F}_{fix}, \sigma(\textbf{ATT}_{fix}) \cdot \textbf{ATT}_{fix}),
\end{aligned}
\end{equation}
where, $\textbf{O}_{mov}$ and $\textbf{O}_{fix}$ are the output feature maps (learned new representations) of $\textbf{F}_{mov}$ and $\textbf{F}_{fix}$, following application of their estimated co-attention maps, respectively. $Softmax(\circ)$ is the Softmax function, applied to the last channel of the similarity matrix $\textbf{S}$. $\sigma(\circ)$ denotes the Sigmoid function, which is a $1\times 1 \times 1$ convolution layer followed by a Sigmoid activation. $Concat(\circ)$ is to concatenate the input feature with the corresponding attention maps, comprising a concatenation, a $1\times 1 \times 1$ convolution layer, a batch-normalisation and an activation layer.

Following the two upsampling blocks in the decoder, the co-attention feature maps of the fixed and moving images are recovered to the original size and resolution of the inputs images. A $3\times3\times3$ convolution followed by a Softmax activation function is used to predict the segmentation masks of the moving and fixed images. The focus in this study is on intra-subject spatio-temporal registration of cine-MR image sequences, i.e. pair-wise registration of images acquired at different time points in the cardiac cycle. We train and evaluate the performance of SDDIR on cardiac cine-MR images available from the UK Biobank imaging database. We focus on segmenting and decomposing the fixed and moving images into four sub-regions, namely, the left ventricle blood pool (LVBP), left ventricle myocardium (LVM), right ventricle (RV) and background. It is important to note that while the focus of this study is on intra-subject cardiac MR image registration, the proposed method is agnostic to imaging modality, organ(s) of interest and application. SDDIR may be used for joint segmentation (into regions/organs of interest) and registration of other types of images (e.g. computed tomography, computed tomography angiography, x-ray, MR angiography, etc.). 

The co-attention block is inspired by~\cite{lu2019see}, but differs in the following ways - (1) Skip-connections are applied in our implementation (as shown in Fig.~\ref{fig:segmentation}), which helps ensure better flow of gradients during training and helps improve overall segmentation performance; (2) In addition to the standard segmentation loss (e.g. Dice loss between the predicted segmentation and ground-truth segmentation), we also compute the cross-entropy between the warped predicted segmentation and the ground-truth fixed segmentation, as a segmentation `consistency' loss, to ensure that the predicted segmentations for the fixed and moving images and the deformation field mapping the latter to the former, are consistent with each other.

\subsection{Discontinuity-preserving Registration}
The segmentation masks predicted for the input fixed and moving images (by the segmentation sub-network) are passed as inputs along with their corresponding original images, to the registration sub-network. To estimate the desired locally smooth and globally discontinuous deformation field mapping the moving image to the fixed image, we first predict four different smooth sub-deformation fields for each of the four sub-regions of interest in the cardiac MR images (i.e. LVBP, LVM, RV and background), and then compose them to obtain the final deformation field. 

\subsubsection{Network Architecture} 
The segmentation masks predicted by the segmentation sub-network are used to split the original pair of fixed and moving images into four different image pairs, comprising, the LVBP, LVM, RV and background sub-regions. As shown in Fig.~\ref{fig:network}, in each pair, the pixel/voxel values within the mask are retained, while those from the surrounding regions are set to zero. Then, a shared-weight U-Net (comprising four downsampling and three upsampling blocks) is used to learn features from all four image pairs. Therefore, we obtain features at $64\times64\times8$ from the original image pairs ($128\times128\times16$). A shared-weight convolution layer followed by a scaling and squaring layer is used to process the learned features and estimate their corresponding diffeomorphic sub-deformation fields. The predicted moving segmentation masks are used again to extract the corresponding regions in the estimated the sub-deformation fields and combine them to obtain the final globally discontinuous deformation field. Finally, a spatial transform network (STN) is used to warp the moving image and predicted segmentation using the composed discontinuous deformation field.

\subsubsection{Discontinuity Composition} 
Composition of deformation fields estimated for relevant image sub-regions is essential to ensure locally smooth and globally discontinuous deformations fields are used for registering images. Similar to previous papers~\cite{dalca2019unsupervised,krebs2019learning,chen2021ddir}, we assume the transformation function (denoted as $\phi_z$) is parameterised by stationary velocity fields (SVF) ($z_i, i \in [0,3]$), which are sampled from a multivariate Gaussian distribution. With the predicted feature map, we obtain four SVFs ($z_0,z_1,z_2,z_3$) corresponding to different regions (LVBP, LVM, RV and background) using a shared-weight convolution layer of size $3\times3\times3$, whose weights are sampled from a Normal distribution. The SVFs are integrated by scaling and squaring layers (referring to~\cite{dalca2019unsupervised}) to diffeomorphic deformation fields. After an upsampling operation, we obtain four diffeomorphic deformation fields $\phi_{z_0}$, $\phi_{z_1}$, $\phi_{z_2}$ and $\phi_{z_3}$. Similarly, we use the predicted moving segmentation masks to extract each region of interest from the obtained deformation fields and compose them to generate the final deformation field. Let the segmentation masks of LVBP, LVM, RV and background be $S_{LVBP}$, $S_{LVM}$, $S_{RV}$  and $S_{B}$ respectively, the composition can be formulated as,
\begin{equation}
\label{eqn:formula_composition}
\phi_z=\phi_{z_0}\times S_{LVBP}+\phi_{z_1}\times S_{LVM}+\phi_{z_2}\times S_{RV}+\phi_{z_3}\times S_{B}.
\end{equation}

\subsection{Loss Function} 
The segmentation and registration sub-networks within the proposed approach are trained jointly using the same loss function ${L}_{total}$. The loss function ${L}_{total}$ includes four terms: segmentation accuracy loss, image similarity loss, segmentation consistency loss and a discontinuity-preserving regularisation term. The segmentation accuracy loss includes two parts, the accuracy loss for the moving image and the fixed image. We use cross-entropy to compute the distance between the predicted segmentation and their respective ground-truth segmentation masks. Denoting the cross-entropy loss as CN, the segmentation accuracy loss ${L}_{seg}$ is formulated as,
\begin{equation}
\label{eqn:seg}
\begin{aligned}
{L}_{seg} = CN(S_{pre}^{mov},S_{gt}^{mov})+CN(S_{pre}^{fix},S_{gt}^{fix}),
\end{aligned}
\end{equation}
where, $S_{pre}^{mov}$, $S_{gt}^{mov}$, $S_{pre}^{fix}$, $S_{gt}^{fix}$ are the predicted and ground-truth segmentations of the moving and fixed images, respectively.

The image similarity loss evaluates the dissimilarity between the warped moving image and the fixed image. In this paper, we use the mean squared error to evaluate the distance between them, formulated as,
\begin{equation}
\label{eqn:mse}
{L}_{MSE} = \frac{1}{W\times H\times D}\sum_{i}^{W\times H\times D}{(I_{mov}\circ \phi - I_{fix})^2}.
\end{equation}

The segmentation consistency loss links the segmentation and registration sub-networks, allowing them to be jointly optimised. This loss term is computed as the Dice overlap~\cite{milletari} between the predicted fixed segmentation and the warped predicted moving segmentation, formulated as,
\begin{equation}
\label{eqn:dice}
{L}_{Dice} = 1- \frac{2|(S_{pre}^{mov} \circ \phi) \cap S_{pre}^{fix}|}{|S_{pre}^{mov} \circ \phi|+|S_{pre}^{fix}|}.
\end{equation}

The discontinuity-preserving regularisation must ensure estimated deformation fields are locally smooth and globally discontinuous. Specifically, dicontinuous at boundaries between structures/regions of interest (i.e. in our case at boundaries between LVBP, LVM, RV and background). Therefore, we cannot enforce a global smoothness constraint on the composed deformation field. As the composition of different deformation fields preserves discontinuities at interfaces, we only need to guarantee the deformation field of each sub-region is smooth. This is achieved by applying $L_2$-regularisation on each sub-deformation field, also referred to as the diffusion regulariser~\cite{dalca2019unsupervised,balakrishnan2019voxelmorph}, denoted $R$, on the spatial gradients of each sub-displacement field $\textbf{u}$. $R$ is formulated as,
\begin{equation}
\label{eqn:regularisation}
\begin{aligned}
&R(\phi) = ||\bigtriangledown \textbf{u}||^2,\\
&L_R = \frac{1}{4}(R_{LVBP} + R_{LVM} + R_{RV} + R_{background}),
\end{aligned}
\end{equation}
where $L_R$ denotes the combined regularisation terms for each sub-region. 

The complete loss function used to train the network is,
\begin{equation}
\label{eqn:total}
\begin{split}
{L}_{total} = \lambda_0 \times {L}_{seg} + \lambda_1 \times L_{MSE}+\lambda_2 \times L_{Dice}+ \lambda_3 \times L_R, 
\end{split}
\end{equation}
where, $\lambda_0$, $\lambda_1$, $\lambda_2$ and $\lambda_3$ are hyper-parameters used to weight the importance of each loss term.

\section{Experiments and Results}
\subsection{Data and Implementation} 
The proposed approach, SDDIR, is trained and evaluated on three publicly available cardiac MR image datasets, namely, the UKBB~\cite{petersen2015uk}, ACDC~\cite{bernard2018deep} and M\&M~\cite{campello2021multi}. We choose 1437 subjects from the UKBB dataset, each including short-axis (SAX) image stacks at ED and ES. The pixel spacing of images in the UKBB are is $\sim1.8 \times 1.8 \times 10 mm^3$. To train the network, we pre-process all image volumes by cropping and padding (with zeros) them to a fixed size of $128\times128\times16$. In this paper, we focus on intra-subject deformable image registration, specifically, to register images from ED to ES and ES to ED. We split the UKBB data into a training set (1080 subjects), a validation set (157 subjects), and test set (200 subjects). This resulted in a total of 2160, 304, and 400 samples (i.e. pair of fixed and moving images for each subject) that were used for training, validation and testing. The ground-truth segmentation masks for the UKBB dataset were manually annotated by experts, as part of a previous study~\cite{petersen2017reference}. To verify the generalisation and robustness of the proposed approach, we also apply the trained model (i.e. trained on UKBB data) to images from the ACDC and M\&M datasets. Similarly, we choose the ED and ES SAX images from 100 subjects (totally 200 samples) in the ACDC dataset, whose ground-truth segmentation masks are available. In the M\&M dataset, 300 samples (registration from ED to ES and from ES to ED) are extracted from 150 subjects. Each image volume in ACDC and M\&M is pre-processed similarly to the UKBB data, resulting in images of size $128\times128\times16$ using resampling, cropping and padding. Note that, to reduce the domain gap between different datasets, histogram-matching is applied to the ACDC and M\&M images, using a random image volume from UKBB as the reference.

The SDDIR was implemented in Python using PyTorch, on a Tesla M60 GPU machine. The Adam optimiser with a learning rate of 1e-3 was used to optimise the network. We set the batch-size to 3, due to limitations in GPU memory available. The hyper-parameters in the total loss $L_{total}$ $\lambda_0$, $\lambda_1$, $\lambda_2$ and $\lambda_3$ were tuned empirically and were set to 0.1, 1, 0.1, 0.01, respectively, throughout all experiments presented in this study. The number of integration steps in the integration layer is 7, following~\cite{dalca2019unsupervised,balakrishnan2019voxelmorph}. The source code will be publicly available on Github (\url{https://github.com/cistib/DDIR}).

\subsection{Competing Methods and Evaluation Metrics} 
To highlight the benefits of the proposed approach, we quantitatively compare the registration performance of SDDIR against both traditional and state of the art deep learning-based registration methods. Three traditional registration methods are compared against SDDIR, namely, the Symmetric Normalisation (SyN, using 3 resolution level, with 100, 80, 60 iterations respectively) in  ANTS~\cite{avants2011reproducible}, Demons (Fast Symmetric Forces Demons~\cite{vercauteren2007diffeomorphic} with 100 iterations and standard deviations 1.0) available in SimpleITK, and B-splines registration (max iteration step is 4000, sampling 4000 random points per iteration), available in SimpleElastix~\cite{marstal2016simpleelastix}. State of the art deep learning-based registration methods chosen for quantitative comparison against the proposed approach include, Voxelmorph (VM~\cite{dalca2019unsupervised}), the weakly-supervised version of VM (denoted as VM-Dice), and a baseline joint segmentation and registration network, named Baseline. VM-Dice essentially trains the original VM approach in a weakly supervised manner, with a Dice loss $L_{Dice}$ on the warped moving segmentation and fixed segmentation (using the ground-truth segmentation masks). We implement the Baseline network by referring to~\cite{xu2019deepatlas,li2019hybrid} based on our setting, which uses a general U-Net for segmentation and a VM like architecture for registration. It is trained with the same loss function as SDDIR, where the only connection between the segmentation sub-network and the registration sub-network is the segmentation consistency loss. All the networks are trained until convergence on the training dataset, and the hyperparameters and final models are selected based on their performance on the validation set.

We also compare the proposed approach, against other discontinuity-preserving registration methods, namely, DDIR~\cite{chen2021ddir} and two other sub-deformation field composition methods investigated previously in~\cite{chen2021ddir}, denoted VM(compose) and VM-Dice(compose). In these composition methods, the original MR images are firstly split into four different pairs, using the ground-truth segmentation masks. Then the trained network (VM or VM-Dice) is used to register those pairs independently. The obtained sub-deformation fields are composed into the final deformation field, which is used to warp the moving image and segmentation. This strategy is a simple and conventional approach to enabling the estimation of discontinuity-preserving deformation fields. In contrast with SDDIR, in this strategy networks are not trained end-to-end and they require segmentation masks to be available during inference.

Registration performance of each method investigated is quantitatively evaluated and compared using the following metrics - Dice scores (computed between the warped moving segmentation and fixed segmentation) on LVBP, LVM and RV, the average Dice score (denoted as Avg. DS) across all cardiac structures, and Hausdorff distance (95\%) (HD95), where, higher Dice score and lower HD95 indicate better registration performance. Additionally, two clinical cardiac indices, the LV end-diastolic volume (LVEDV) and LV myocardial mass (LVMM), are also computed to demonstrate that the proposed registration approach preserves clinically relevant volumetric indices post image registration. They are calculated based on the moving segmentation and the corresponding warped moving segmentation (for example, using the ED segmentation and warped ED segmentation when registering ED to ES). The closer clinical indices to the reference (i.e. the clinical indices computed based on ground-truth ED and ES segmentation, presented in the row `before Reg'), the better. To assess segmentation performance, the average Dice scores (denoted as Seg DS) across all cardiac structures for the predicted moving and fixed segmentation masks, with respect to their corresponding ground-truth segmentation masks, are also calculated.

\begin{table*}[ht] 
 \caption{\label{tab:comparison} Quantitative comparison on UKBB between SDDIR and state of the art methods. Statistically significant improvements in registration and segmentation accuracy (DS and HD95) are highlighted in bold. Besides, LVEDV and LVMM indices with no significant difference from the reference (before Reg) are also highlighted in bold.}
 \centering
 \resizebox{12cm}{!}{ 
 \begin{tabular}{lcccccccc} 
 \hline
  Methods  & Avg. DS (\%) & LVBP DS (\%) & LVM DS (\%) & RV DS (\%) & HD95 (mm) & Seg DS (\%)& LVEDV & LVMM \\
 \hline
  before Reg & $43.75 \pm 5.47$ & $63.14 \pm 14.17$ & $52.93 \pm 13.14$ & $68.12 \pm 15.36$ & $15.44 \pm 4.56$ &-& $157.53 \pm 32.78$ & $99.50 \pm 27.89$\\ 
  B-spline & $66.23 \pm 6.70$ & $74.26 \pm 9.54$ & $62.12 \pm 8.08$ & $62.31 \pm 7.88$ & $15.92 \pm 4.92$ & -& $140.34 \pm 41.13$ & $\textbf{100.27 $\pm$ 30.25}$ \\ 
  Demons & $68.29 \pm 6.03$ & $77.51 \pm 8.32$ & $63.79 \pm 7.17$ & $63.56 \pm 7.58$ & $14.44 \pm 4.58$ & -& $139.64 \pm 37.30$ & $\textbf{102.53 $\pm$ 30.26}$\\
  SyN & $54.78 \pm 5.60$ & $65.15 \pm 6.38$ & $41.09 \pm 8.34$ & $58.11 \pm 6.89$ & $16.10 \pm 4.82$  & - & $150.19 \pm 33.11$ & $93.44 \pm 28.22$\\ 
  \hline
  VM & $74.18 \pm 4.96$ & $85.50 \pm 6.38$ & $69.43 \pm 6.55$ & $67.61 \pm 7.44$ & $12.74 \pm 4.60$ & -&$151.85 \pm 33.76$ & $\textbf{97.08 $\pm$ 29.25}$ \\
  VM-Dice & $79.80 \pm 4.42$ & $86.96 \pm 5.80$ & $71.72 \pm 6.77$ & $80.71\pm 5.90$ & $8.67 \pm 4.56$&-& $\textbf{162.09 $\pm$ 34.40}$ & $\textbf{100.36 $\pm$ 28.57}$\\ 
  Baseline & $79.95 \pm 4.38$ & $88.33 \pm 5.08$ & $73.61 \pm 6.28$ & $77.91\pm 6.58$ & $9.51 \pm 4.25$&$87.31 \pm 2.95$& $\textbf{156.93 $\pm$ 33.64}$ & $\textbf{98.46 $\pm$ 28.53}$\\ 
 \hline
 VM(compose) & $69.77 \pm 5.72$ & $75.77 \pm 5.57$ & $57.50 \pm 9.59$ & $76.03 \pm 6.38$ & $10.06 \pm 2.88$ & -&$128.89 \pm 41.42$ & $\textbf{100.54 $\pm$ 30.34}$\\ 
 VM-Dice(compose) & $65.95 \pm 6.93$ & $74.36 \pm 7.26$ & $52.66 \pm 11.13$ & $70.84 \pm 6.62$ & $11.10 \pm 2.98$ &-& $124.06 \pm 44.59$ & $\textbf{102.38 $\pm$ 29.35}$\\ 
 \hline
 DDIR & $86.54 \pm 4.30$  & $90.47\pm 4.88 $ & $80.62 \pm 6.20 $ & $88.53 \pm 5.14$ & $6.59 \pm 4.37$ &-& $\textbf{155.65 $\pm$ 32.92}$ & $\textbf{101.65 $\pm$ 27.87}$ \\
 SDDIR & $84.46 \pm 3.64$ & $90.75\pm 3.84$ & $78.07\pm 6.48$ & $84.56 \pm 5.21$ & $7.54 \pm 4.83$ &$88.60 \pm 2.06$ & $\textbf{156.02 $\pm$ 32.66}$ & $\textbf{101.28 $\pm$ 27.90}$\\ 
 SDDIR(-DC) & $80.41 \pm 4.18$ & $88.41\pm 5.03$ & $74.24\pm 6.29$ & $78.58 \pm 6.46$ & $9.44 \pm 4.10$ &$88.07 \pm 2.44$ & $\textbf{155.53 $\pm$ 33.50}$ & $\textbf{98.43 $\pm$ 28.44}$\\
 \hline
 \end{tabular}
  }
\end{table*}

\begin{figure*}[ht]
\begin{center}
\includegraphics[width=\textwidth]{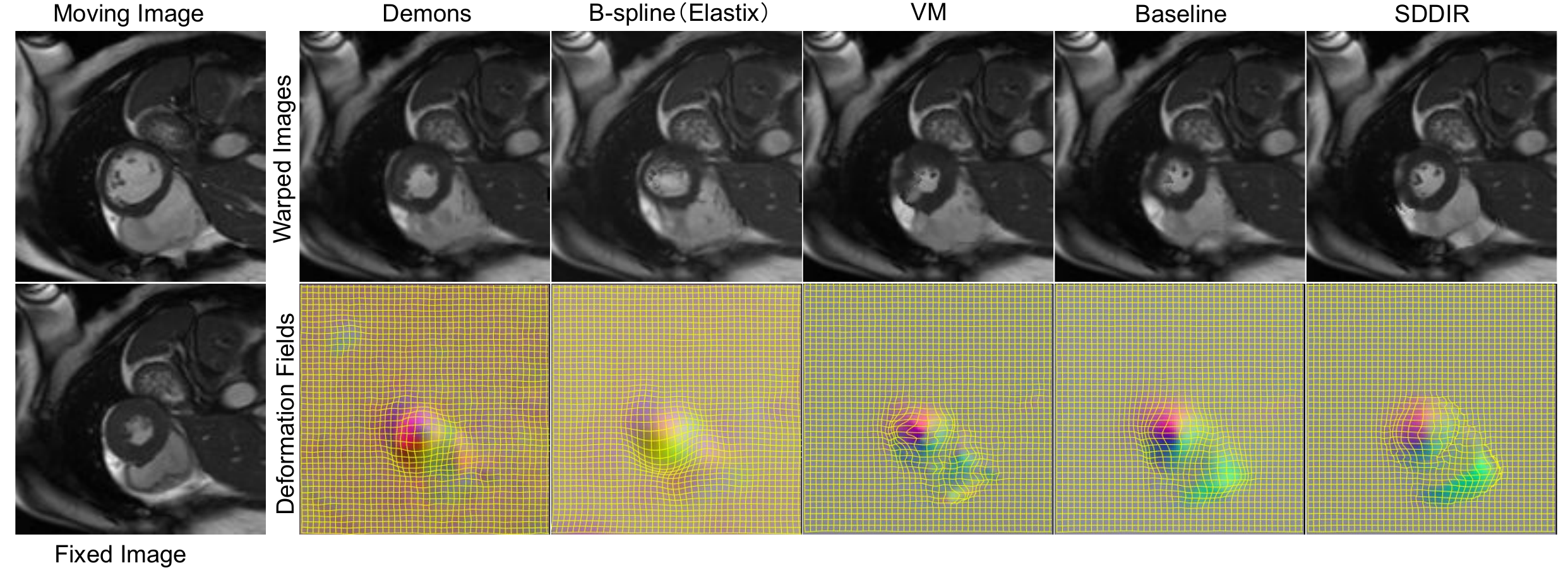}
\caption{Visual comparison of results on UKBB estimated using SDDIR and state of the art methods. Left column: moving and fixed images; Right column: corresponding warped moving image (first row), deformation fields (second row). The cardiac MR images were reproduced by kind permission of UK Biobank ©.}
\label{fig:UKBB_qualitative_results}
\end{center}
\end{figure*}

\subsection{Registration Results: UKBB Data} 
Quantitative registration results obtained for the unseen test set from UKBB are summarised in Table~\ref{tab:comparison}. The Baseline joint segmentation and registration network achieves higher Dice scores on the registration results than those solely designed for image registration (e.g. Demons, B-spline and VM). The composition methods VM(compose) and VM-Dice(compose) do not show any improvements over VM and VM-Dice, and perform consistently worse than VM-Dice across all metrics evaluated. While the Baseline network significant outperforms VM across all metrics, its average Dice score (computed across all three cardiac structures, LVBP, LVM and RV), only marginally outperforms VM-Dice, and it performs worse than VM-Dice in terms of HD95. Both the Baseline and VM-Dice networks use the Dice loss to guide the training of their constituent registration networks. The architectures of the constituent registration networks are almost identical, leading to similar performance of the Baseline and VM-Dice networks. SDDIR significantly outperforms the Baseline network in terms of both the Dice score and HD95, highlighting the superior registration performance of the proposed approach. The DDIR approach, which uses manually annotated ground truth segmentation masks for decomposing the input images into regions of interest during inference, achieves the best registration performance of all methods investidated. DDIR achieves an average Dice score 5\% higher than SDDIR and an HD95 score that is on average 2 mm lower than SDDIR. In terms of the clinical indices derived from the registered/warped moving image following registration, both DDIR and SDDIR show no significant differences with respect to the reference (with indices derived for SDDIR being closer to the reference). While DDIR shows some improvements over SDDIR in terms of registration accuracy, the latter does not require segmentation masks to be provided as inputs during inference (as required by DDIR). SDDIR thus has more flexibility in its utility/application and is better suited to real-world scenarios where segmentation masks for regions of interest may not be available prior to registering the input images.

Registration results visualised in Fig.~\ref{fig:UKBB_qualitative_results} indicate that the warped moving image predicted by SDDIR is more similar to the fixed image than those predicted by the other methods investigated, which is consistent with the quantitative results obtained. This is especially evident along the boundaries of the right ventricle and the left ventricular myocardium. Additionally, Fig.~\ref{fig:UKBB_qualitative_results} indicates that compared with all other approaches, the deformation field estimated by SDDIR captures discontinuities at boundaries between different structures/sub-regions (such as between the left and right ventricle, for example) more strongly.

\subsection{Registration Results: ACDC and M\&M Data} 
To assess the ability of the proposed approach to generalise to unseen data representative of real-world data acquired routinely in clinical examinations, we apply the SDDIR model pre-trained on UKBB data, to other external cardiac MR data sets, namely, ACDC and M\&M. Data available in ACDC and M\&M were acquired at multiple different imaging centres distributed across different countries, using different types of MR scanners, and from patients diagnosed with different types of cardiac diseases/abnormalities (e.g. myocardial infarction, dilated cardiomyopathy, hypertrophic cardiomyopathy, abnormal right ventricle). Generalising to such unseen data is challenging due to domain shifts in the acquired images, relative to the UKBB data used for training SDDIR.

\begin{table*}[ht] 
 \caption{\label{tab:comparison_acdc} Quantitative comparison on ACDC between SDDIR and the state of the art methods. Statistically significant improvements in registration accuracy (DS and HD95) are highlighted in bold. Besides, LVEDV and LVMM indices with no significant difference from the reference are also highlighted in bold.}
 \centering
 \resizebox{12cm}{!}{ 
 \begin{tabular}{lcccccccc} 
 \hline
  Methods  & Avg. DS (\%) & LVBP DS (\%) & LVM DS (\%) & RV DS (\%) & HD95 (mm) & Seg DS (\%)& LVEDV & LVMM \\
 \hline
  before Reg & $60.24 \pm 11.19$ & $65.70 \pm 16.22$ & $51.90 \pm 14.50$ & $63.14 \pm 14.17$ & $10.60 \pm 3.88$ &-& $165.13 \pm 73.57$ & $130.09 \pm 50.58$\\ 
  B-spline & $74.96 \pm 9.55$ & $80.40 \pm 14.16$ & $75.72 \pm 7.84$ & $68.75 \pm 15.50$ & $10.96 \pm 4.66$ & -& $\textbf{152.52 $\pm$ 82.11}$& $\textbf{134.36 $\pm$ 51.37}$ \\ 
  Demons & $73.54 \pm 8.83$ & $78.42 \pm 12.58$ & $72.34 \pm 8.90$ & $69.87 \pm 13.79$ & $10.52 \pm 3.92$ & -& $147.02 \pm 81.01$ & $\textbf{137.06 $\pm$ 53.38}$ \\ 
  SyN & $70.03 \pm 7.58$ & $79.66 \pm 10.56$ & $66.08 \pm 8.59$ & $64.35 \pm 14.69$ & $10.56 \pm 3.47$  & - & $\textbf{155.95 $\pm$ 75.71}$ & $\textbf{133.03 $\pm$ 51.74}$\\ 
  VM & $76.02 \pm 7.84$ & $83.85 \pm 10.42$ & $74.15 \pm 8.25$ & $70.07 \pm 13.99$ & $9.85 \pm 4.36$ & -&$\textbf{156.62 $\pm$ 75.93}$ & $\textbf{130.54 $\pm$ 51.92}$ \\ 
  VM-Dice & $77.75\pm 7.40$ & $84.67 \pm 9.92$ & $73.86 \pm 8.24$ & $74.71\pm 12.39$ & $8.03 \pm 3.91$&-& $\textbf{164.70 $\pm$ 77.20}$ & $\textbf{134.75 $\pm$ 54.01}$\\ 
  Baseline & $78.64 \pm 6.76$ & $86.31 \pm 8.64$ & $76.60 \pm 7.42$ & $73.02\pm 12.55$ & $9.19 \pm 4.00$ & $66.40 \pm 18.39$& $\textbf{159.99 $\pm$ 74.93}$ & $\textbf{132.48 $\pm$ 52.60}$\\ 
 \hline
 VM(compose)& $72.57 \pm 10.13$ & $76.85 \pm 14.08$ & $64.03 \pm 13.67$ & $76.84 \pm 12.45$ & $9.73 \pm 2.99$ & -&$137.96 \pm 82.08$ & $\textbf{129.18 $\pm$ 54.00}$\\ 
 VM-Dice(compose)& $71.01 \pm 9.22$ & $77.29 \pm 11.92$ & $62.54 \pm 12.63$ & $73.21 \pm 12.71$ & $10.03 \pm 3.24$ &-& $140.07 \pm 81.06$ & $\textbf{130.72 $\pm$ 53.44}$\\ 
 \hline
 DDIR & $87.79 \pm 5.56$  & $91.53\pm 6.62 $ & $85.49 \pm 7.16 $ & $86.34 \pm 9.13$ & $5.86 \pm 4.54$ &-& $\textbf{161.03 $\pm$ 74.98}$ & $\textbf{134.96 $\pm$ 53.08}$ \\
 SDDIR & $76.13 \pm 8.66$ & $83.75\pm 12.59$ & $73.22\pm 10.38$ & $71.42 \pm 13.10$ & $9.38 \pm 3.50$ &$\textbf{72.64 $\pm$ 15.17}$ & $\textbf{155.40 $\pm$ 73.17}$ & $\textbf{134.31 $\pm$ 57.69}$\\
 SDDIR(-DC) & $79.61 \pm 6.67$ & $86.97\pm 8.52$ & $77.73\pm 6.91$ & $74.12 \pm 12.49$ & $8.84 \pm 3.90$ &$70.10 \pm 13.86$ & $\textbf{160.57 $\pm$ 75.90}$ & $\textbf{132.40 $\pm$ 52.52}$\\
 \hline
 \end{tabular}
  }
\end{table*}

\begin{figure*}[ht]
\begin{center}
\includegraphics[width=\textwidth]{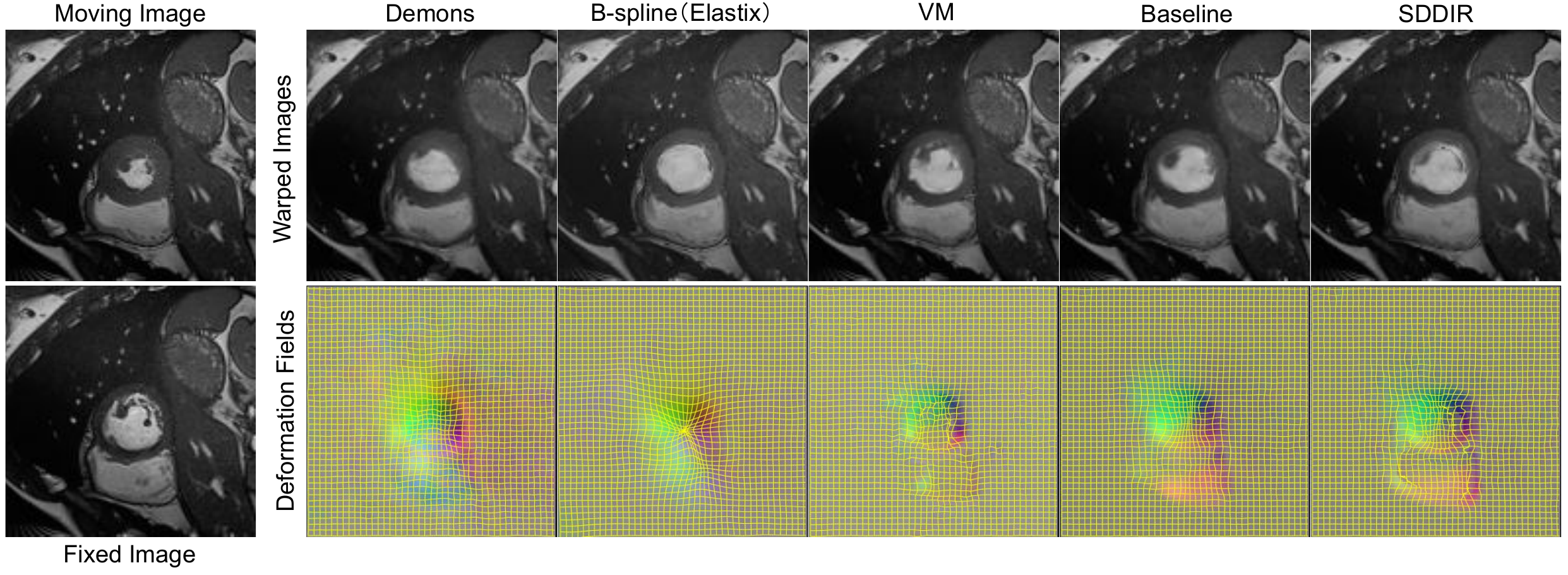}
\caption{Visual comparison of results on ACDC estimated using SDDIR and state of the art methods. Left column: moving and fixed images; Right column: corresponding warped moving image (first row), deformation fields (second row).}
\label{fig:ACDC_qualitative_results}
\end{center}
\end{figure*}

\begin{table*}[ht] 
 \caption{\label{tab:comparison_mm} Quantitative comparison on MM between SDDIR and the state of the art methods. Statistically significant improvements in registration and segmentation accuracy (DS and HD95) are highlighted in bold. Besides, LVEDV and LVMM indices with no significant difference from the reference are also highlighted in bold.}
 \centering
 \resizebox{12cm}{!}{ 
 \begin{tabular}{lcccccccc} 
 \hline
  Methods  & Avg. DS (\%) & LVBP DS (\%) & LVM DS (\%) & RV DS (\%) & HD95 (mm) & Seg DS (\%)& LVEDV & LVMM \\
 \hline
  before Reg & $54.13 \pm 10.40$ & $60.36 \pm 12.47$ & $44.30 \pm 14.23$ & $57.75 \pm 10.61$ & $15.04 \pm 4.86$ &-& $161.43 \pm 63.86$ & $124.37 \pm 46.15$\\ 
  B-spline & $65.57 \pm 10.87$ & $70.65 \pm 14.84$& $67.03 \pm 9.29$ & $59.05 \pm 13.26$ & $18.77 \pm 7.24$ & -& $144.61 \pm 72.60$& $121.49 \pm 44.76$ \\ 
  Demons & $65.76 \pm 10.02$ & $71.74 \pm 12.97$ & $65.04 \pm 9.71$ & $60.49 \pm 12.46$ & $16.86 \pm 6.67$ & -& $142.32 \pm 69.74$ & $\textbf{125.15 $\pm$ 45.97}$ \\ 
  SyN & $58.03 \pm 10.05$ & $66.45 \pm 13.68$ & $58.18 \pm 9.93$ & $49.47 \pm 13.05$ & $17.33 \pm 5.25$  & - & $138.46 \pm 69.53$ & $116.48 \pm 44.10$\\ 
  VM & $68.91 \pm 8.19$ & $76.65 \pm 10.03$ & $67.91 \pm 8.61$ & $62.16 \pm 11.31$ & $15.70 \pm 5.91$ & -&$144.81 \pm 65.96$ & $115.38 \pm 45.30$ \\
  VM-Dice & $72.69\pm 7.55$ & $79.33 \pm 9.26$ & $67.60 \pm 7.92$ & $71.14\pm 11.50$ & $11.80 \pm 4.04$&-& $\textbf{164.66 $\pm$ 66.18}$ & $\textbf{125.61 $\pm$ 46.18}$\\ 
  Baseline & $73.69 \pm 7.48$ & $81.36 \pm 9.04$ & $70.81 \pm 7.48$ & $68.91\pm 11.10$ & $12.48 \pm 4.21$ & $60.44 \pm 22.08$& $\textbf{154.36 $\pm$ 65.33}$ & $\textbf{117.78 $\pm$ 44.23}$\\ 
 \hline
 VM(compose)& $69.36 \pm 9.67$ & $75.28 \pm 10.72$ & $57.77 \pm 13.94$ & $75.03 \pm 10.94$ & $10.81 \pm 2.87$ & -& $132.27 \pm 71.60$&  $\textbf{117.57 $\pm$ 46.56}$\\ 
 VM-Dice(compose)& $66.65 \pm 9.82$ & $74.18 \pm 11.55$ & $56.10 \pm 12.85$ & $69.66 \pm 11.70$ & $13.22 \pm 5.79$ &-& $133.66 \pm 70.65$ & $\textbf{120.54 $\pm$ 46.34}$\\
 \hline
 DDIR & $86.11 \pm 5.71$  & $89.37\pm 6.14 $ & $81.50 \pm 6.41 $ & $87.47 \pm 8.65$ & $6.85 \pm 4.90$ &-& $\textbf{154.01 $\pm$ 64.57}$ & $\textbf{128.00 $\pm$ 46.19}$ \\
 SDDIR & $69.02 \pm 10.32$ & $76.78\pm 12.81$ & $65.74\pm 10.93$ & $64.55 \pm 13.69$ & $12.78 \pm 4.06$ &$\textbf{66.11 $\pm$ 16.65}$ & $\textbf{153.04 $\pm$ 64.29}$ & $\textbf{122.69 $\pm$ 46.99}$\\
 SDDIR(-DC) & $74.44 \pm 7.65$ & $81.80\pm 9.21$ & $71.66\pm 7.37$ & $69.86 \pm 11.40$ & $12.24 \pm 4.38$ &$63.64 \pm 15.39$ & $\textbf{155.06 $\pm$ 64.14}$ & $\textbf{119.12 $\pm$ 44.79}$\\
 \hline
 \end{tabular}
  }
\end{table*}

The quantitative and qualitative results obtained for data from ACDC are shown in Table~\ref{tab:comparison_acdc} and Fig.~\ref{fig:ACDC_qualitative_results}, respectively. For the results on ACDC we observed that B-spline and Demons achieve better registration performance than VM, while the SyN algorithm is still worse than VM (by $\sim 2\%$ in terms of the average Dice score). The VM-Dice and Baseline networks are significantly better than VM and two composition-based image registration approaches, namely, VM (compose) and VM Dice (compose). As Baseline only uses the predicted segmentations to compute the Dice loss (rather than using it to partition the original images as in SDDIR), registration quality of the Baseline network is less dependent on segmentation quality. As a result, although the segmentation performance of Baseline is significantly worse, it performs comparably to SDDIR in terms of registration quality on ACDC, achieving 1\% higher Dice, and a marginally lower average HD95 score. Using ground-truth segmentation masks during inference, DDIR performs the best out of all models investigating, achieving an average Dice score of 87\% Dice score. The registration performance of SDDIR drops significantly relative to the results obtained for UKBB data, with SDDIR performing marginally worse than the Baseline and VM-Dice networks in terms of the Dice and HD95 metrics (compared with Baseline, no significant difference on HD95 and RV Dice, while average Dice, LV Dice and LVM Dice significantly decreased) used to evaluate registration performance (see columns 2-6 in Table~\ref{tab:comparison_acdc}). Conversely, SDDIR obtains $>6\%$ improvement in segmentation accuracy, evaluated using the Dice score, relative to the Baseline network. This is mainly because the registration sub-network in SDDIR is highly-dependent on the segmentation sub-network, to split the original MR images into pairs of sub-regions. Consequently, segmentation errors are propagated to the subsequent registration step and overall registration performance drops significantly when predicted segmentations are of poor quality. The clinical indices predicted by most methods, except Demons and the two composition methods, show no significant differences to the reference. Student's t-test was used throughout to assess statistical significance, considering a significance threshold of $5\%$, (i.e. p-values are $>0.05$). To our analysis, this is because the cardiac motion in ACDC is not so large as UKBB (as shown in Fig.~\ref{fig:ACDC_qualitative_results}), leading to less difference in the anatomical structures even when the registration is not good enough. As shown in Fig.~\ref{fig:ACDC_qualitative_results}, the Demons and B-spline tend to predict warped moving images with over-smoothed image features and object boundaries, losing local details (e.g. the papillary muscles in the left ventricle). The deep learning-based registration methods obtain more such localised anatomical details more consistently. 

The quantitative results on the M\&M dataset are shown in Table~\ref{tab:comparison_mm}. Similar to the results on ACDC, traditional methods B-spline and Demons outperform VM, while SyN still performs worse than VM. VM-Dice and Baseline obtain higher average Dice scores than the traditional methods and the composition methods. DDIR obtains the highest registration performance (86.11\% on average Dice score), while the performance of SDDIR is significantly decreased to 69.02\%, due to poor performance of the segmentation sub-network (66.11\%). Despite the drop in registration performance, SDDIR predicts registered/warped images that show no statistically significant differences from the reference in terms of relevant clinical indices (LVEDV, LVMM), similar to DDIR, VM-Dice and the Baseline network (Student's t-test, p-value $>0.05$). 

\subsection{Segmentation Analysis} 
\begin{figure*}[!ht]
\begin{center}
\includegraphics[width=\textwidth]{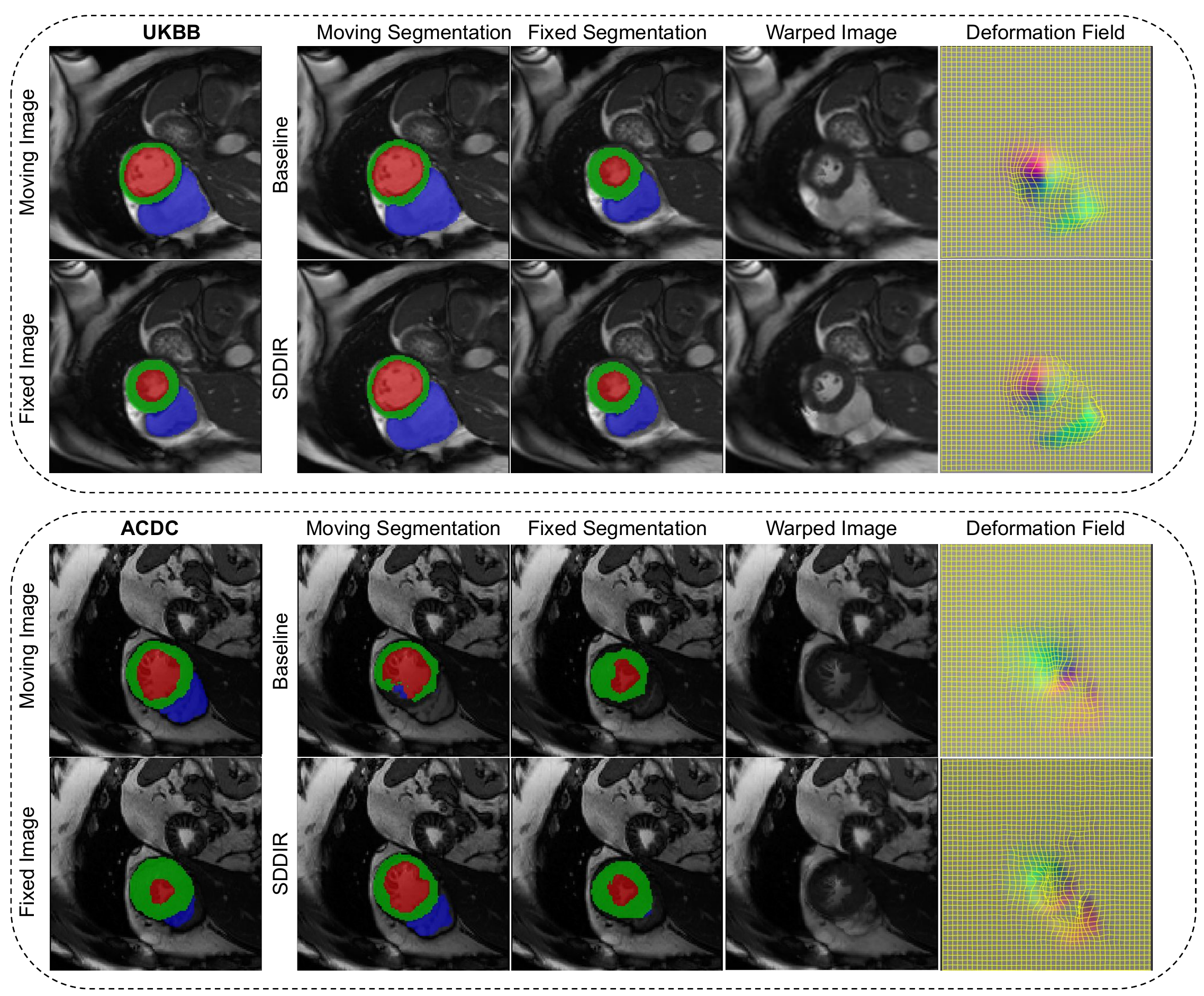}
\caption{Visual comparison of segmentation estimated using SDDIR and Baseline. The first two rows are results on UKBB (green dashed border) and the bottom two rows (red dashed border) are the results on ACDC. Left column: moving and fixed images with the corresponding segmentation; Right column: corresponding moving segmentation, fixed segmentation, warped moving image, deformation fields. The cardiac MR images were reproduced by kind permission of UK Biobank ©.}
\label{fig:segmentation_results}
\end{center}
\end{figure*}

In this section, we analyse on the segmentation performance of SDDIR and the Baseline network. Examples of segmentation masks predicted using either approach are shown in Fig.~\ref{fig:segmentation_results}, and quantitative results summarising the segmentation accuracy of both approaches are presented in Table~\ref{tab:comparison}, Table~\ref{tab:comparison_acdc} and Table~\ref{tab:comparison_mm}. The segmentation results obtained for the UKBB, summarised in Table~\ref{tab:comparison}, show that SDDIR achieves $1\%$ higher Dice score than the Baseline network across the unseen test data. This improvement in segmentation accuracy is more pronounced for the ACDC and M\&M datasets, with SDDIR achieving $\sim$ 6\% improvement in the Dice score for both datasets, relative to the Baseline network (72.64\% vs 66.40\% and 66.11\% vs 60.44\%). These results demonstrate that SDDIR significantly outperforms the Baseline network in terms segmentation accuracy (for the input fixed and moving images), consistently across multiple datasets. 

Fig.~\ref{fig:segmentation_results} highlights the ability of both the Baseline network and SDDIR to predict high quality segmentation masks for the input image pairs (fixed and moving) from the UKBB dataset. Their performance on data from ACDC however, degrades, producing masks that are visually different to the ground-truth (e.g. the segmentation of the right ventricle, plotted in blue in Fig.~\ref{fig:segmentation_results}). In the UKBB dataset, the results of SDDIR are more similar to the ground-truth, but there is over-segmentation in the results of Baseline (see the region of the right ventricle in the moving segmentation). There is a domain gap from the UKBB to ACDC dataset, which leads to the decreased segmentation Dice for both methods. However, SDDIR offers some improvements over the Baseline, for example, by capturing the right ventricle in cases where it is entirely missed by the latter.  In summary, due to the co-attention block, SDDIR is more robust (than the Baseline) for segmenting the input pairs of images in the presence of domain shifts.

\subsection{Ablation Study} 
To analyse the contribution of each block in SDDIR, we conducted an ablation study on the proposed network, using UKBB data. The results are shown in Table~\ref{tab:ablation}, where, SDDIR(-DC), SDDIR(-CA), SDDIR(-Seg) and SDDIR(-Reg) denote removing the discontinuity composition block, co-attention block, segmentation sub-network and registration sub-network in SDDIR, respectively. By comparing these variants of SDDIR, we can assess the contribution of the proposed joint segmentation and registration sub-networks, co-attention block and discontinuity-preserving strategy. Without the discontinuity composition block, the SDDIR(-DC) is essentially a globally smooth registration method. SDDIR(-CA) applies the same segmentation sub-network as Baseline, whilst still ensuring discontinuity-preserving registration. By removing the segmentation sub-network, the SDDIR(-Seg) turns into a registration network similar to VM. Correspondingly, the SDDIR(-Reg) is a sole co-attention based segmentation network.

\begin{table*}[ht] 
 \caption{\label{tab:ablation} Quantitative comparison on UKBB between different versions of SDDIR. Statistically significant improvements in registration and segmentation accuracy (DS and HD95) are highlighted in bold. Besides, LVEDV and LVMM indices with no significant difference from the reference are also highlighted in bold.}
 \centering
 \resizebox{12cm}{!}{ 
 \begin{tabular}{lcccccccc} 
 \hline
  Methods  & Avg. DS (\%) & LVBP DS (\%) & LVM DS (\%) & RV DS (\%) & HD95 (mm) & Seg DS (\%)& LVEDV & LVMM \\
 \hline
  before Reg & $43.75 \pm 5.47$ & $63.14 \pm 14.17$ & $52.93 \pm 13.14$ & $68.12 \pm 15.36$ & $15.44 \pm 4.56$ &-& $157.53 \pm 32.78$ & $99.50 \pm 27.89$\\ 
  SDDIR(-DC) & $80.41 \pm 4.18$ & $88.41 \pm 5.03$ & $74.24 \pm 6.29$ & $78.58 \pm 6.46$& $9.45 \pm 4.10$ &$88.07 \pm 2.44$& $\textbf{155.53 $\pm$ 33.50}$ & $\textbf{98.43 $\pm$ 28.44}$\\
  SDDIR(-CA) & $75.71 \pm 5.32$ & $82.49 \pm 7.82$ & $70.12 \pm 7.12$ & $74.51 \pm 7.13$ & $9.55 \pm 4.14$ & $84.94 \pm 3.77$&$149.26 \pm 33.54$ & $108.19 \pm 31.85$\\ 
  SDDIR(-Reg) & - & - & - & - & -&$87.38 \pm 4.31$& - & -\\ 
  SDDIR(-Seg) & $74.20 \pm 4.80$ & $85.82 \pm 6.11$ & $69.59 \pm 6.45$ & $67.17 \pm 7.50$ & $12.85 \pm 4.64$ & -&$\textbf{153.01 $\pm$ 33.78}$ & $93.32 \pm 136.10$ \\
 \hline
 SDDIR & $\textbf{81.56 $\pm$ 4.15}$ & $88.65\pm 6.40$ & $75.48\pm 6.37$ & $\textbf{80.54 $\pm$ 5.95}$ & $\textbf{8.63 $\pm$ 4.41}$ &$88.38 \pm 2.03$ & $\textbf{156.02 $\pm$ 32.66}$ & $\textbf{101.28 $\pm$ 27.90}$\\
 \hline
 \end{tabular}
  }
\end{table*}

According to Table~\ref{tab:ablation}, without the co-attention block in the segmentation sub-network, the segmentation and registration performance of SDDIR(-CA) is significantly lower than SDDIR. After removing the discontinuity composition block, the average registration Dice score of SDDIR(-DC) is significantly decreased, as it is unable to ensure globally discontinuous and locally smooth deformation fields. Comparing SDDIR with the corresponding variants of the network that tackle purely segmentation (SDDIR(-Reg)) and registration tasks (SDDIR(-Seg)), we find that the joint segmentation and registration framework improves the performance of each sub-network for each corresponding task. This indicates that the two tasks are mutually beneficial to each other.

Although the overall registration performance of SDDIR is worse than DDIR, the former outperforms other state of the art methods, and ensures that cardiac clinical indices derived from the warped/registered images show no statistically significant differences to the reference (derived from the original images). Furthermore, SDDIR does not require high-quality segmentation masks to be available a priori for the input images to be registered, unlike DDIR (which was trained and evaluated using segmentation masks delineated manually by experts). SDDIR thus lends itself to use real clinical applications where high-quality segmentation masks are seldom available, and has the added benefit of producing good quality segmentation masks for both input images to be registered, as auxilliary outputs. Although the globally smooth version of SDDIR (i.e. without discontinuity composition), SDDIR(-DC), incurs significantly higher registration errors than SDDIR for the UKBB data set, it consistently outperforms other state-of-the-art approaches in terms of registration accuracy and joint registration and segmentation performance, across all three datasets. Considering those scenarios where globally smooth registration is required/appropriate, SDDIR(-DC) can be employed in place of SDDIR to reduce dependency of registration accuracy on segmentation quality, and improve overall registration performance. For example, results reported for ACDC and M\&M data sets in Tables \ref{tab:comparison_acdc}, \ref{tab:comparison_mm} show that SDDIR(-DC) outperforms SDDIR in terms of registration accuracy, at the cost of enforcing global smoothness on the estimated deformation fields.

\section{Discussion and Conclusion}
The proposed approach, SDDIR, is versatile and can be employed in various clinical applications requiring pair-wise image registration. For example, the ability to simultaneously segment and register cardiac MR images means that SDDIR can facilitate real-time quantification of cardiac clinical indices across the cardiac cycle and quantitative analysis of cardiac motion. In addition, as SDDIR can predict both fixed segmentation and warped moving segmentation, a more anatomical structure-plausible segmentation results can be obtained by joint considering those two predictions. SDDIR is agnostic to image modality and the organs/structures visible within the field of view of the images being registered. Hence, SDDIR may also be used to jointly segment and register thoracic or abdominal CT images, where strong discontinuities exist between organ structures due to their relative motion (i.e. sliding at organ boundaries) resulting from respiration. 

Although the proposed approach is demonstrated to jointly segment and register input pairs of images accurately, outperforming several state-of-the-art approaches, one main limitation remains. The registration performance of SDDIR is highly dependent on the performance of the segmentation sub-network, i.e. on the quality of the segmentation masks predicted for the input pair of images to be registered. As the registration sub-network requires the predicted segmentation masks to split the original MR images into pairs of corresponding regions, the registration sub-network performs poorly when the quality of the segmentation masks is poor. Thus, SDDIR performs well on the UKBB dataset as accurate segmentation masks are predicted for the input pairs of images and used to effectively guide the discontinuity-preserving registration. As SDDIR was trained using UKBB data, the segmentation sub-network was able to generalise well to unseen data from UKBB due to homogeneity/consistency in appearance across images from different subjects. Conversely, SDDIR's performance significantly degraded when the trained model (on UKBB data) was used to register images from other datasets (e.g. ACDC and MM). This is due to domain shifts resulting from variations in imaging scanners and protocols, used to acquire images in different datasets. The presented results indicate that SDDIR outperforms the Baseline network in terms of registration accuracy only when good quality segmentation masks are predicted by its constituent segmentation sub-network. Specifically, we found that segmentation accuracy (in terms of Dice) on the fixed and moving images had to be over 76\% for ACDC and 70\% for MM, for the subsequent registration accuracy of SDDIR to be better than the Baseline network. Therefore, future work in the field should look to improve the robustness of discontinuity-preserving image registration methods to domain shifts that are commonly found in medical images. This may be achieved by imbuing SDDIR with recent approaches to domain generalisation, for example, to mitigate for the drop in segmentation and registration performance resulting from domain shifts (relative to the training data). Additionally, the over-dependence of registration quality on the quality of the segmentation masks predicted by SDDIR could be relaxed by modelling object/tissue boundaries as weak discontinuities (as opposed to strong discontinuities used currently in SDDIR) that are incorporated into the regularisation of the deformation field to ensure locally smooth and globally discontinuous deformation fields. 

In this paper, we propose a novel weakly-supervised discontinuity-preserving registration network, SDDIR. The proposed approach is applied to the task of intra-patient spatio-temporal CMR registration, to jointly segment the input pair of images to be registered and predict a locally smooth but globally discontinuous deformation field that warps the source/moving image to the fixed/target image. Compared with previous discontinuity-preserving registration methods, SDDIR provides improvements in terms of execution speed (relative to traditional iterative approaches), and does not required segmentation masks to be available prior to registering images (unlike some state-of-the-art deep learning based registration approaches such as DDIR). We demonstrate the registration performance of SDDIR on three cardiac MR datasets, and prove that it can significantly outperform both traditional and deep learning-based state-of-the-art registration methods. Future works will explore domain generalisation techniques to mitigate for the drop in performance observed with SDDIR due to domain shifts and will look to weaken the dependency on segmentation quality to ensure accurate image registration.

\section*{Acknowledgements}
This research was conducted using the UKBB resource under access application 11350 and was supported by the Royal Academy of Engineering under the RAEng Chair in Emerging Technologies (CiET1919/19) scheme.

\bibliographystyle{splncs04} 
\bibliography{sddir}
\end{document}